\documentclass[runningheads]{llncs}
\usepackage[T1]{fontenc}
\usepackage{graphicx}
\usepackage{soul}
\usepackage{url}
\usepackage{hyperref}

\usepackage{verbatim}
\usepackage{subfigure}
\usepackage{calc}
\usepackage{amstext,amsmath}
\usepackage{stackengine}
\newcommand\IncG[2][]{\addstackgap{%
\raisebox{-.5\height}{\includegraphics[#1]{#2}}}}
\usepackage{array}
\usepackage{amsmath,amssymb} 
\usepackage{xcolor}
\usepackage{paralist}
\usepackage{multirow}

\usepackage{algorithm}  
\usepackage{algpseudocode}

\begin{document}
%

\title{Boosting LightWeight Depth Estimation Via Knowledge Distillation}
%
%
\author{Junjie Hu\inst{1}\orcidID{0000-0002-1911-4361} \and
 Chenyou Fan \inst{2} \and
Hualie Jiang\inst{3} \and Xiyue Guo\inst{4} \and Yuan Gao\inst{1} \and Xiangyong Lu\inst{5} and Tin Lun Lam \inst{3,1}\orcidID{0000-0002-6363-1446}}
\authorrunning{J.Hu et al.}
%
\institute{Shenzhen Institute of Artificial Intelligence and Robotics for Society, China \and South China Normal University, China \and
The Chinese University of Hong Kong, Shenzhen, China \and Zhejiang University, China \and Tohoku University, Japan
\\
\email{T.L.Lam is the corresponding author, tllam@cuhk.edu.cn}}
%
\maketitle              
\begin{abstract}

Monocular depth estimation (MDE) methods are often either too computationally expensive or not accurate enough due to the trade-off between model complexity and inference performance. In this paper, we propose a lightweight network that can accurately estimate depth maps using minimal computing resources. We achieve this by designing a compact model architecture that maximally reduces model complexity.
To improve the performance of our lightweight network, we adopt knowledge distillation (KD) techniques. We consider a large network as an expert teacher that accurately estimates depth maps on the target domain. The student, which is the lightweight network, is then trained to mimic the teacher's predictions. However, this KD process can be challenging and insufficient due to the large model capacity gap between the teacher and the student.
To address this, we propose to use auxiliary unlabeled data to guide KD, enabling the student to better learn from the teacher's predictions. This approach helps fill the gap between the teacher and the student, resulting in improved data-driven learning. Our extensive experiments show that our method achieves comparable performance to state-of-the-art methods while using only 1\% of their parameters. Furthermore, our method outperforms previous lightweight methods regarding inference accuracy, computational efficiency, and generalizability. Code is available on \url{https://github.com/JunjH/Boosting-Light-Weight-Depth-Estimation}.

\end{abstract}

\section{Introduction}

Monocular depth estimation has gained widespread attention as an economical and convenient alternative to depth sensors, providing applications in obstacle avoidance \cite{Mancini2016FastRM}, simultaneous localization and mapping (SLAM) \cite{Tateno2017CNNSLAMRD,Guo2020SemanticHB}, robot navigation \cite{mendes2020deep}.
With the rapid development of deep learning in recent years, significant progress has been made in this field.

\begin{figure}[t]
\centering
\includegraphics[width=0.72\textwidth]{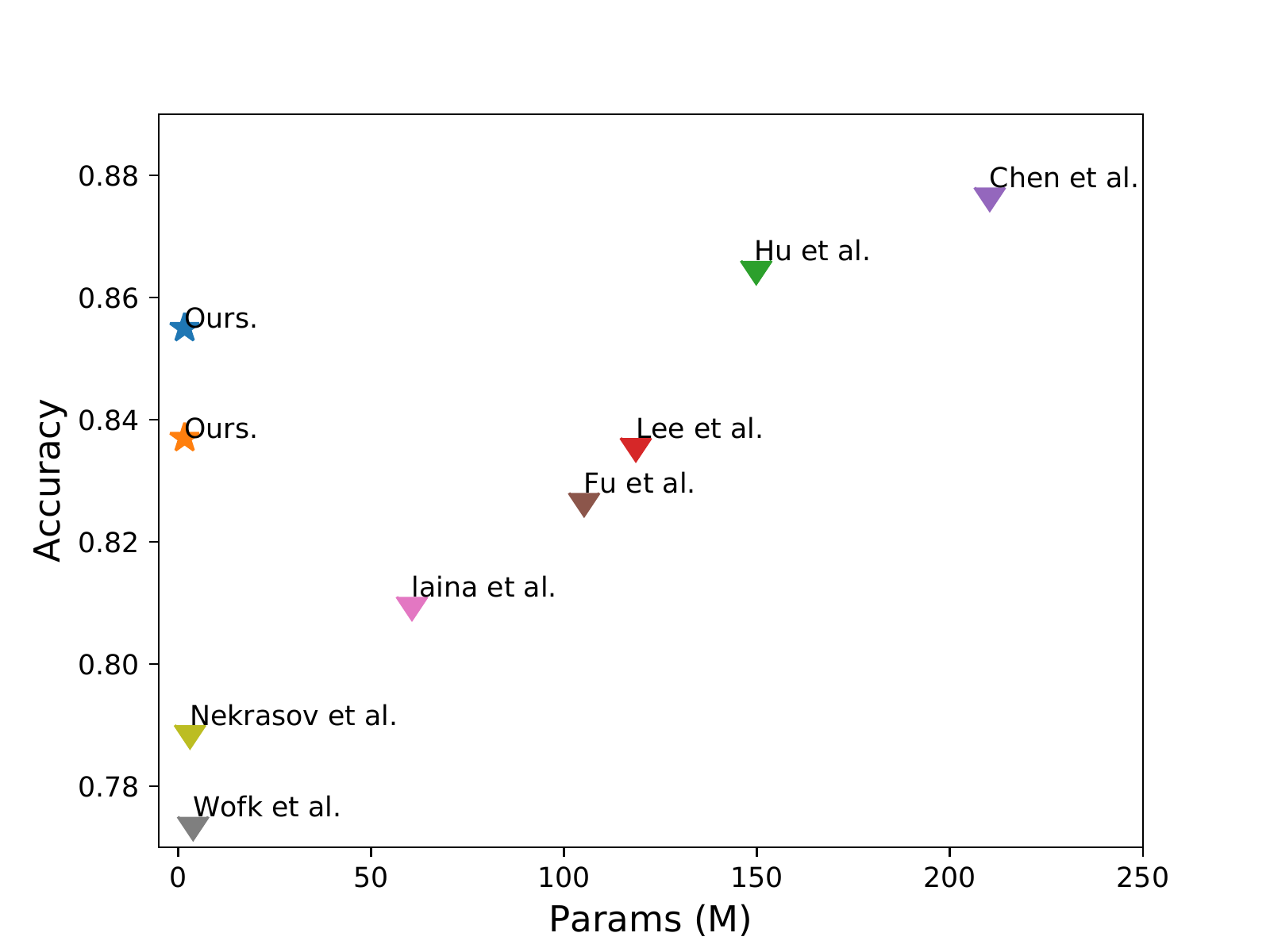}
\vspace{-5mm}
\caption{The total parameters and accuracy of different methods for depth estimation. As seen, there is a trade-off between accuracy and model complexity. Our method achieves competitive performance compared with state-of-the-art methods with just a tiny portion (1$\%$) of parameters.}
\label{fig_config}
\vspace{-5mm}
\end{figure}

Most of the previous works mainly focused on the improvement of estimation accuracy \cite{fu2018deep,laina2016deeper,hu2019revisiting}.
However, the depth estimation has to be both computationally efficient and accurate.
It is essential for real-world applications with limited computation resources.
Although several prior works have attempted to improve the computational efficiency with lightweight networks \cite{Wofk2019FastDepthFM,Nekrasov2019RealTimeJS}, they often come at the cost of significantly decreased inference accuracy. 
There is an urgent need for MDE to achieve satisfactory performance while maintaining good efficiency.

In this paper, we propose a novel approach to monocular depth estimation that aims to achieve high inference accuracy with minimal hardware resources. To achieve this goal, we introduce a lightweight network with a compact architecture design that reduces the model complexity while maximizing the accuracy of the depth estimation. Unlike traditional encoder-decoder architectures, our approach compresses the feature maps extracted by multi-layers of the encoder to a fixed number of channels at each scale and then upsample them to the same resolution. These feature maps are then concatenated and fed to two convolutional layers to produce the final depth map. Our network, built on MobileNet-v2, has only 1.7M parameters, making it one of the most lightweight networks in the literature. By minimizing the model complexity, we aim to strike a balance between accuracy and computational efficiency, making our approach well-suited for real-world applications with limited hardware resources.

\begin{figure*}[t]
\centering
\includegraphics[width=\linewidth]{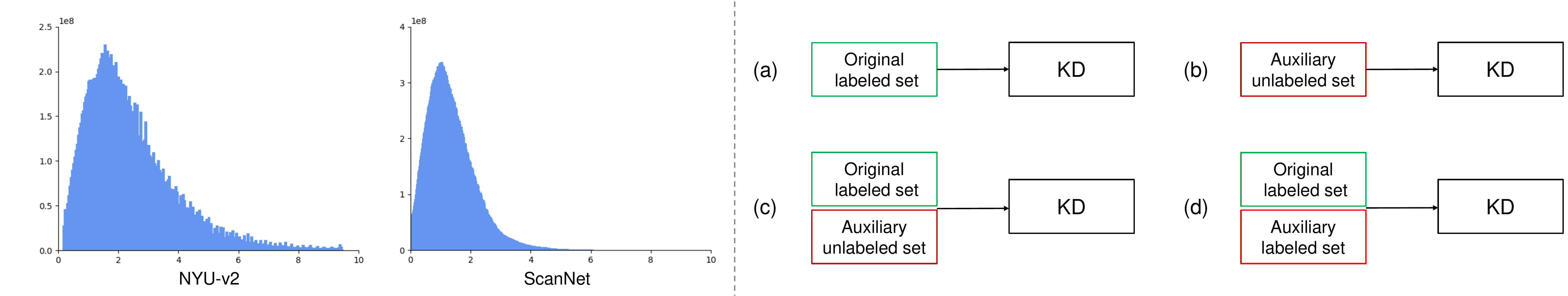}
\vspace{-5mm}
\caption{The left side shows the depth histogram of NYU-v2 training set and ScanNet validation set, respectively. Both of the two histograms exhibit a long-tailed distribution and they are highly similar. The right side shows the configuration of knowledge distillation (KD) considered in this paper. 
(a) is the standard method that applies KD with the original labeled set. (b) applies KD with only the auxiliary unlabeled set. (c) applies KD with the original labeled set and the auxiliary unlabeled set. (d) applies KD with both the original and auxiliary labeled set.
}
\label{fig_frame}
\vspace{-4mm}
\end{figure*}

We next describe our approach to training the lightweight network using knowledge distillation (KD) \cite{Romero2015FitNetsHF}. Specifically, we leverage a large network trained on the target domain $\mathcal{X}$ to serve as an expert teacher. Given any input image from $\mathcal{X}$, the teacher network outputs the corresponding depth map.
To improve the performance of the lightweight network, we propose a novel approach to promote KD using auxiliary data. Our approach is motivated by two considerations. First, since depth estimation is a non-linear mapping from RGB space to depth space, KD can be seen as an approximation of this mapping in a data-driven way. Therefore, the more high-quality data we have, the more accurately we can approximate the mapping. Second, in depth estimation, we find that auxiliary data can be more easily collected since many real-world scenarios share similar scene scales and demonstrate similar depth histograms.
For example, two popular indoor benchmarks, NYU-v2 and ScanNet, exhibit similar long-tailed depth distributions and depth ranges, as shown in Fig.~\ref{fig_frame}. This observation motivates our proposal to use auxiliary data to guide KD, which enables the lightweight network to leverage additional training signals and improve its accuracy.

Our study focuses on two scenarios: labeled and unlabeled data. In the labeled case, ground truths can be obtained using depth sensors such as Kinect. In the unlabeled case, auxiliary data can be collected using a visual camera in scenarios with similar scene scales. Therefore, leveraging auxiliary data is practical for improving the performance of depth estimation in real-world applications.
In this paper, we propose to take the following learning strategies for these two specific cases. To this end, we propose the following learning strategies for these two cases:
\begin{itemize}
\item When auxiliary data is unlabeled, we first train the teacher on the original labeled set and then apply KD with both the original labeled set and the auxiliary unlabeled set to improve the student.

\item When auxiliary data is labeled, we first train the teacher on the combined original and auxiliary sets, which provides a more discriminative teacher. We then apply KD to further enhance the student with the mixed dataset.
\end{itemize}

Using auxiliary data can effectively improve the performance of lightweight depth estimation, as demonstrated in Fig.~\ref{fig_config}. Our proposed method achieves comparable results to state-of-the-art methods, while utilizing only $1\%$ of the parameters, and outperforms other lightweight approaches by a large margin. To evaluate the effectiveness of our approach, we conduct a series of experiments and confirm that:
\begin{itemize}
\item  Even without access to the original training set, our approach can still be effective if enough auxiliary unlabeled samples are available and they have similar scene scales to the original training samples.

\item Combining the original trained set and auxiliary unlabeled set in KD can significantly improve performance by better bridging the gap between the teacher and student.

\item Directly training the lightweight network with a mixed dataset of both original and auxiliary labeled data has limited improvement due to its low capacity. However, the two-stage learning strategy of first training a larger teacher and then applying KD is more effective in this case.
    

\end{itemize}

\section{Related Work}

\subsection{Monocular Depth Estimation}
In previous studies, monocular depth estimation has been addressed in a supervised learning approach by minimizing the pixel-wise loss between the predicted and ground truth depth \cite{Eigen2014depth,hu2019revisiting}. Various network architectures have been proposed, including the basic encoder-decoder network \cite{laina2016deeper}, networks with skip connections \cite{hu2019revisiting}, dilated convolution \cite{fu2018deep}, and pyramid pooling \cite{Mendes2021OnDL}, all of which have shown improved performance. Additionally, the problem can be formulated as an unsupervised learning task, where the geometry consistency of multi-view images is taken into account \cite{Zhou2017UnsupervisedLO}. However, the performance of unsupervised approaches still lags behind supervised methods.

Real-time depth estimation has also been investigated in several studies. For example, lightweight networks based on MobileNet and MobileNet-v2 were introduced for fast depth estimation in \cite{Wofk2019FastDepthFM,Nekrasov2019RealTimeJS}, using traditional supervised learning methods. Additionally, an unsupervised approach for depth estimation was proposed in \cite{Liu2020MiniNetAE} using a lightweight network with recurrent modules. While these small networks demonstrate superior computation speed, their accuracy tends to be significantly lower compared to larger networks.

\subsection{Learning with Auxiliary Data}
To improve the performance of learning-based methods, it has become increasingly popular to leverage additional labeled training datasets. This strategy has been shown to be effective in image recognition on ImageNet, with approaches that use extra data achieving a top-1 accuracy that is greater than 5\% higher than methods that do not utilize additional data \cite{pham2020meta,foret2020sharpness}.

In the field of depth estimation, this strategy has also been employed. For example, Chen et al. \cite{Chen2020ImprovingMD} used six auxiliary datasets to handle challenging scenarios such as low light, reflective surfaces, and spurious edges, resulting in superior performance for indoor depth estimation. Additionally, some methods have used multi-domain datasets, such as indoor and outdoor scenes, as well as synthesized and real-world images, to learn a universal network \cite{romanov2020towards,Lasinger2020TowardsRM,yin2020diversedepth,Aleotti2021RealTimeSI}. However, these methods often do not account for the scale differences across datasets, and can only estimate a normalized depth map. In contrast, in this paper, we focus specifically on indoor depth estimation and aim to reconstruct the true scale of the scene.

It is worth noting that auxiliary unlabeled data is also commonly leveraged in image recognition under the setting of semi-supervised learning \cite{Wang2021KnowledgeDA,Xie2020SelfTrainingWN}. In image recognition, auxiliary data must be carefully collected such that its semantic attribute corresponds to the model's predicted categories. In the case of depth estimation, however, auxiliary data is easier to collect in practical scenarios, as it can be obtained simply by capturing additional images with a visual camera.

Overall, leveraging additional labeled and unlabeled data is a promising strategy for improving the performance of depth estimation methods.

\subsection{Knowledge Distillation}
In recent years, knowledge distillation has been extensively studied, originally proposed to transfer knowledge from a large teacher model to a smaller student model in image recognition \cite{Hinton2015DistillingTK}. However, recent studies have attempted to improve the effectiveness of knowledge distillation. Mirzadeh et al. proposed using an assistant network between the teacher and student \cite{Mirzadeh2020ImprovedKD}, while other works have employed intermediate features to guide student learning \cite{Huang2017LikeWY,Liu2020StructuredKD}. The strategy of distilling from multiple teachers has also been proposed \cite{Tarvainen2017MeanTA}. Additionally, some methods augment the training set using techniques such as GANs \cite{Snelson2005SparseGP} or leveraging extra data on the cloud \cite{Xu2019PositiveUnlabeledCO}.

While a few works have applied knowledge distillation to depth estimation \cite{Pilzer2019RefineAD,Aleotti2021RealTimeSI}, it is unclear how auxiliary data can improve knowledge distillation for this task. In contrast, we propose utilizing auxiliary labeled and unlabeled data to improve knowledge distillation for depth estimation, based on the observation that many scenarios have similar scene scales in the real world.



\section{lightweight Network}
 \begin{figure*}[t]
    \centering
    \includegraphics[width=\linewidth]{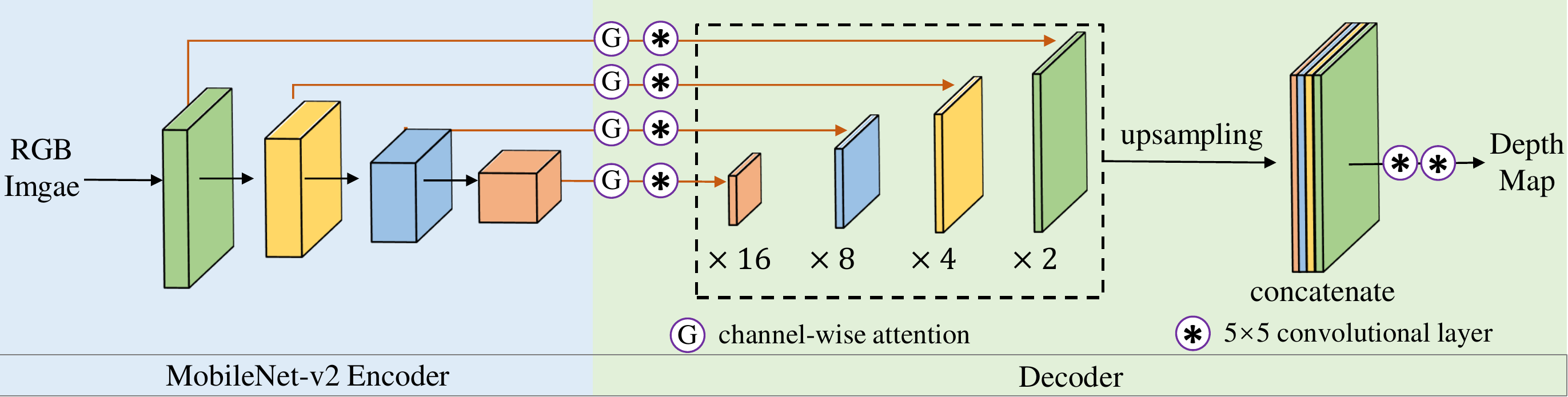}
    \vspace{-5mm}
    \caption{Diagram of the proposed lightweight network.} 
    \label{fig_arch}
    \vspace{-5mm}
\end{figure*}

Most previous works for pixel to pixel regression tasks use a symmetric encoder-decoder network \cite{laina2016deeper,ma2017sparse,Hu2020ATU}. However, these networks can be computationally inefficient, requiring significant GPU memory during computation. Furthermore, research on CNNs has shown that there is a high degree of redundancy within them, with multiple filters capturing similar feature representations \cite{netdissect2017}. To improve the efficiency of depth estimation networks, we propose an extremely compact network architecture in this paper.

\begin{table}[!t]
\begin{center} 
\caption{The details of model parameters for the teacher and student.}
\label{net_paras}
\begin{tabular}{l|cc}
\hline
Network & Teacher & Student \\ \hline
 Backbone & ResNet-34 & MobileNet-v2\\
 Encoder (M)  & 21.3 &1.6\\
 Decoder (M)  & 1.4 &0.3 \\
 Total (M) &21.9  &1.7\\
\hline
\end{tabular}
\end{center}
\vspace{-5mm}
\end{table}

The network architecture we propose is depicted in Fig.\ref{fig_arch} and is based on a lightweight design that achieves high inference efficiency. 
Specifically, given a set of feature maps extracted by encoder blocks, we first apply channel-wise attention \cite{hu2018senet} to attribute weights to each feature map. We then fuse them using the convolutional layer and compress them to a fixed number of channels (16 channels) to reduce the model's complexity. For features extracted with an encoder at multiple scales, we apply the above operation at each scale, and the outputted feature maps are upsampled by factors of $\times2$, $\times4$, $\times8$, and $\times16$, respectively. Finally, we concatenate them and feed them into two $5 \times 5$ convolutional layers to obtain the final depth map.

We adopt ResNet-34 and MobileNet-v2 as backbone networks for the teacher and student, respectively, resulting in  21.9 M and 1.7 M parameters, respectively. The detailed information is given in Table.~\ref{net_paras}.

\section{Promoting KD with Auxiliary Data}

\subsubsection{Standard KD}

We adopt the classical knowledge distillation framework, which involves a well pre-trained teacher network on a labeled set $\mathcal{X}$. The student network is trained using the ground truth depths and estimations from the teacher network as supervision. We denote the teacher and student networks as $N_t$ and $N_s$, respectively. The loss function used to train the student network is defined as follows:

\begin{equation}
\mathcal{L} = \frac{1}{|\mathcal{X}|} \sum_{x_i, g_i \in \mathcal{X}} (\lambda L (N_s(x_i), N_t(x_i)) + (1- \lambda) L (N_s(x_i), g_i) )
\end{equation}

Here, $g_i$ represents the ground truth depth for input $x_i$, $L$ is an error measure between two depth maps, and $\lambda$ is a hyperparameter that balances the two loss terms. To compute $L$, we use the error measure proposed in \cite{hu2019revisiting}, which takes into account depth, gradient, and normal losses.

\subsubsection{Learning with Auxiliary Data}

\begin{figure}[t]
\centering
\subfigure {\includegraphics[width=0.58\linewidth]{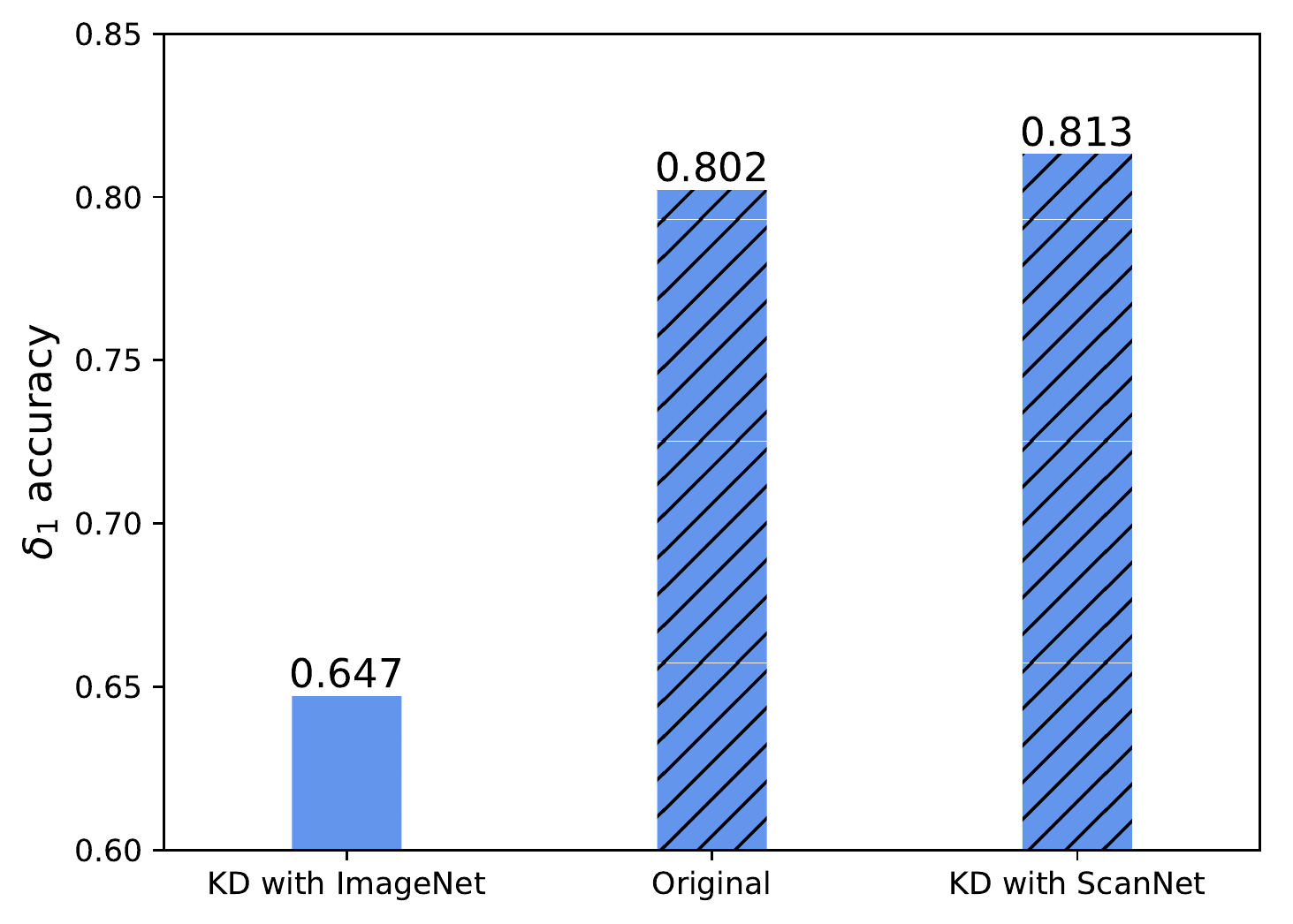}}
\vspace{-3mm}
\caption{Results of the student network on NYU-v2 test set. From left to right: the result of the student that learned via KD with the training set of ImageNet, the result with supervised learning on NYU-v2 training set, and the result of the student that learned via KD with the training set of ScanNet.}
\label{fig_ax_res}
\vspace{-5mm}
\end{figure}

 We hypothesize that auxiliary data can be effective for knowledge distillation (KD) in depth estimation, as long as it shares similar scene scales. To verify this assumption, we conducted a preliminary experiment using a teacher network trained on the NYU-v2 dataset and performing KD with cross-domain datasets.

The loss function used for training the student is defined as:
\begin{equation}
\mathcal{L} = \frac{1}{\mathcal{U}} \sum_{u_j \in \mathcal{U}} \lambda L(N_s(u_j), N_t(u_j))
\end{equation}
where $\mathcal{U}$ denotes the unlabeled set, $N_s$ and $N_t$ are the student and teacher networks, respectively, and $L$ is the error measure between two depth maps.

To evaluate the effectiveness of using auxiliary data, we selected two datasets with different characteristics: ImageNet, an out-of-distribution dataset, and ScanNet, another indoor dataset with similar scene scales. Note that only RGB images from these datasets were used.

As shown in Fig.~\ref{fig_ax_res}, our experiment demonstrated that using ScanNet as auxiliary data resulted in slightly better performance than the original training data alone, while using ImageNet led to a $15.5\%$ accuracy drop. These findings confirm our hypothesis that KD with unlabeled data is effective for depth estimation, provided that the data has similar scene scales to the original data.

Moreover, we found that incorporating both the original training data and auxiliary data further improves the performance of the lightweight network. We considered two scenarios for using auxiliary data, which are discussed in detail in Sec.~\ref{experimental_details}.

\textbf{The use of auxiliary unlabeled data:}
The teacher network, denoted as $N_t$, is trained on the original labeled set $\mathcal{X}$. During the knowledge distillation process, we have access to both $\mathcal{X}$ and an auxiliary unlabeled set $\mathcal{U}$. The loss function used to train the student network is formulated as follows:
\begin{equation}
\begin{split}
    \mathcal{L} = & 
    \frac{1}{\mathcal{X}} \sum_{x_i, g_i \in \mathcal{X}} (\lambda  L (N_s(x_i), N_t(x_i))  + (1- \lambda) L (N_s(x_i),g_i) ) + \\ 
    &  \frac{1}{\mathcal{U}} \sum_{u_j \in \mathcal{U}}  (L (N_s(u_j), N_t(u_j))
\end{split}    
\end{equation}

\textbf{The use of auxiliary labeled data:}
In this case, the auxiliary data $\mathcal{U}^{'}$ is fully labeled, which means that we have access to both the input images and their corresponding ground truth depth maps. We use this data to train a teacher network, denoted as $N_t'$, on a mixed dataset, i.e., $\mathcal{X} \cup \mathcal{U}^{'}$. Since the teacher network is trained on a larger and more diverse dataset, it is expected to be more discriminative than the one trained on $\mathcal{X}$ only.

Next, we use the teacher network $N_t'$ to perform KD on a student network $N_s$, which is learned using both the labeled set $\mathcal{X}$ and the auxiliary labeled set $\mathcal{U}'$. The loss for the student is formulated as:
\begin{equation}
\begin{split}
    \mathcal{L} = & 
    \frac{1}{\mathcal{X}} \sum_{x_i, g_i \in \mathcal{X}} (\lambda  L (N_s(x_i), N'_t(x_i))  + (1- \lambda) L (N_s(x_i),g_i) ) + \\ 
    &  \frac{1}{\mathcal{U}^{'}} \sum_{u_j, g_j \in \mathcal{U}^{'}}  (\lambda L (N_s(u_j), N'_t(u_j))  + (1- \lambda) L (N_s(u_j),g'_j) )
\end{split}    
\end{equation}
where $g'_j$ denotes ground truth of $u_j$.


\section{Experiments}



\subsection{Experimental Setting}

We conducted all experiments on the NYU-v2 dataset \cite{Silberman2012IndoorSA}, which is widely used in previous studies and contains various indoor scenes. We followed the standard preprocessing procedure \cite{Eigen2014depth,laina2016deeper,ma2017sparse}. Specifically, we used the official splits of 464 scenes, with 249 scenes for training and 215 scenes for testing. This resulted in approximately 50,000 unique pairs of images and depth maps with a size of 640$\times$480 pixels. To reduce the computational complexity, we resized the images down to 320$\times$240 pixels using bilinear interpolation and then cropped their central parts to 304$\times$228 pixels, which served as inputs to the networks. The depth maps were resized to $152\times 114$ pixels. For testing, we used the same small subset of 654 samples as in previous studies.

To obtain auxiliary data, we randomly selected 204,000 images from 1,513 scenarios of the ScanNet dataset \cite{dai2017scannet}.

\subsubsection{Implementation Details}
\label{experimental_details}
We adopt ResNet-34 as the teacher network and MobileNet-v2 as the student network. Both networks are trained for 20 epochs, and the loss weight $\lambda$ is set to 0.1. We initialize the encoder module in the network with a model pre-trained on the ImageNet dataset \cite{deng2009imagenet}, while the other layers are initialized randomly. We employ the Adam optimizer with an initial learning rate of 0.0001, a weight decay of 0.0001, and $\beta_{1}=0.9$ and $\beta_{2}=0.999$. We reduce the learning rate to 10$\%$ for every 5 epochs.

\begin{figure*}[!t]
\centering
\subfigure []{\includegraphics[width=0.45\linewidth]{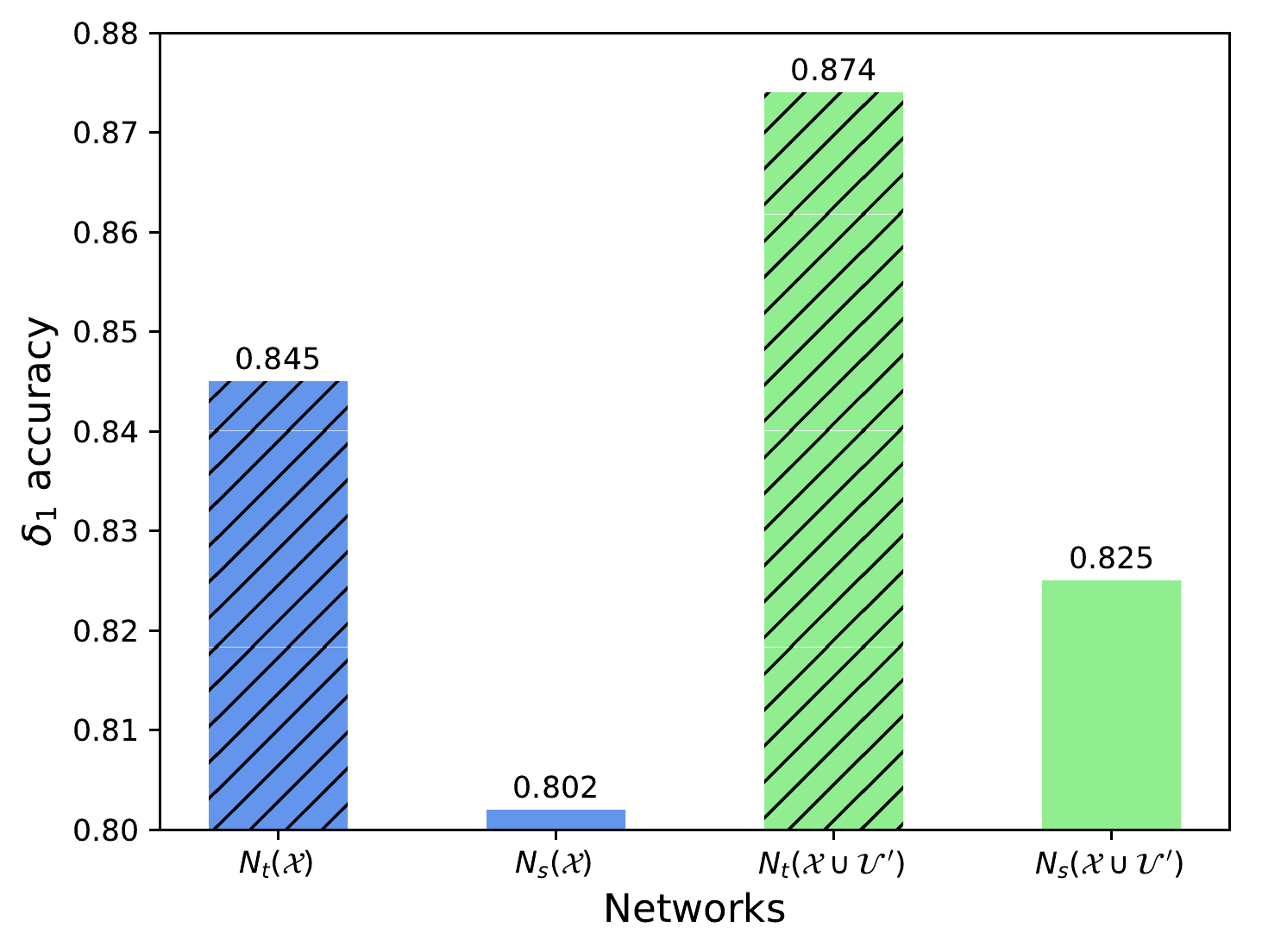}}\quad
\subfigure []{ \includegraphics[width=0.45\linewidth]{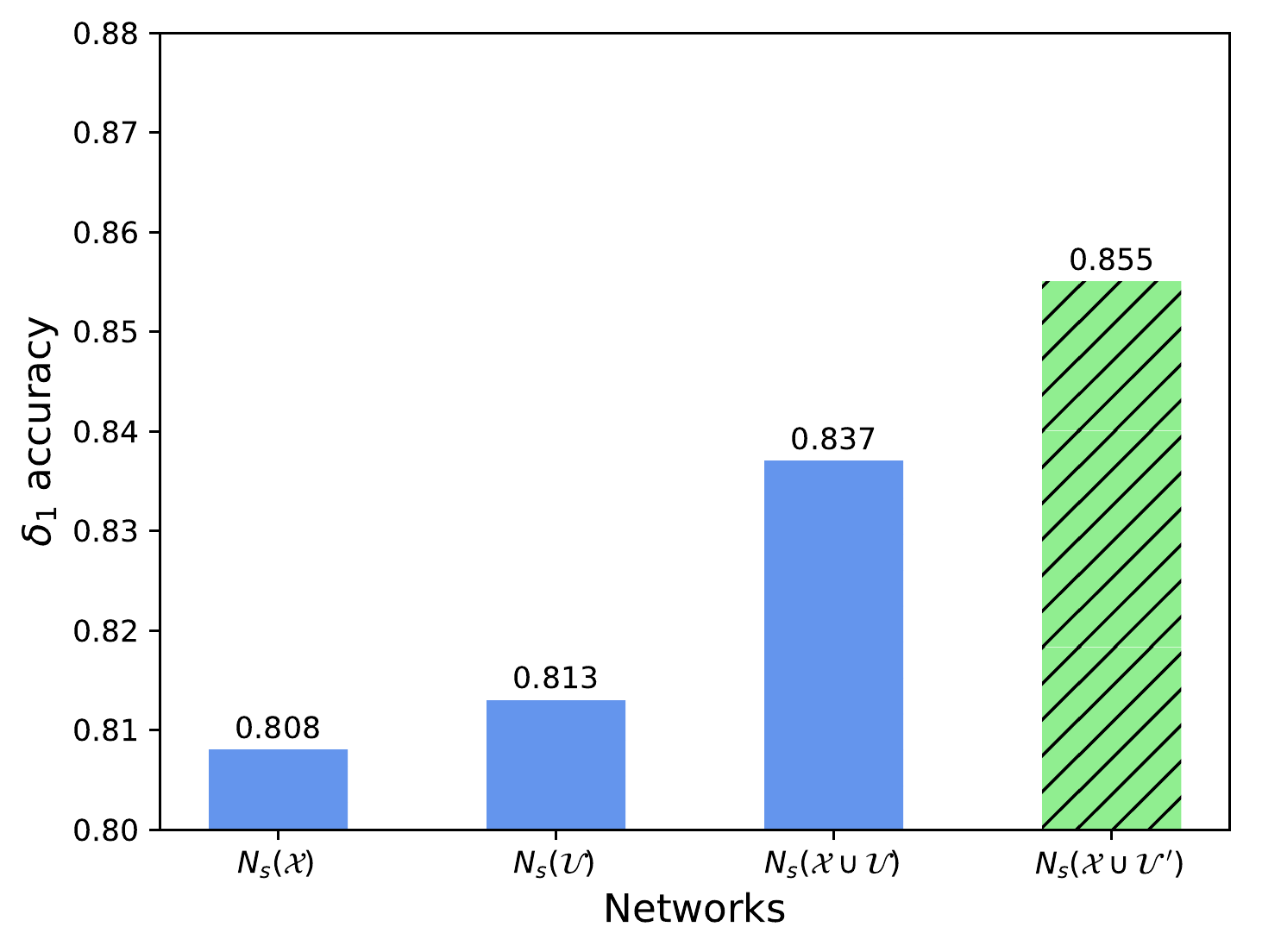}}
\vspace{-3mm}
\caption{(a) Results of the teacher and student network trained with supervised learning. The blue color denotes results trained on $\mathcal{X}$ and the green color denotes results with $\mathcal{X}\cup \mathcal{U'}$.  (b) Results of the student network learned with KD. The blue color denotes results of  $N_s(\mathcal{X})$, $N_s(\mathcal{U})$ and $N_s(\mathcal{X \cup U})$, respectively, and the green color denotes results of $N_s(\mathcal{X \cup U'})$.}
\label{w_wo_kd}
\vspace{-5mm}
\end{figure*}

\subsection{Quantitative Evaluation}
To simplify our notation, we use the following conventions throughout the paper. Specifically, we use $\mathcal{X}$, $\mathcal{U}$, and $\mathcal{U'}$ to refer to the NYU-v2 dataset, the unlabeled ScanNet dataset, and the labeled ScanNet dataset, respectively. The teacher models trained on $\mathcal{X}$ and $\mathcal{X} \cup \mathcal{U}'$ are denoted as $N_t(\mathcal{X})$ and $N_t(\mathcal{X} \cup \mathcal{U}')$, respectively. Similarly, the student models trained on $\mathcal{X}$, $\mathcal{X} \cup \mathcal{U}$, and $\mathcal{X} \cup \mathcal{U}^{'}$ are denoted as $N_s(\mathcal{X})$, $N_s(\mathcal{X} \cup \mathcal{U})$, and $N_s(\mathcal{X} \cup \mathcal{U}^{'})$, respectively.

\subsubsection{Performance without KD}
We first evaluate the teacher and student network with supervised learning. We perform experiments on  $\mathcal{X}$ and the mixed dataset $\mathcal{X}\cup\mathcal{U'}$, respectively. The results are shown in Fig.~\ref{w_wo_kd} (a).
It can be observed that increasing the amount of labeled data leads to performance improvements for both the teacher and student networks, with the teacher improving from 0.845 to 0.874 and the student improving from 0.802 to 0.825. However, a significant performance gap still exists between the teacher and student networks, with the teacher outperforming the student, e.g., 0.845 vs 0.802 and 0.874 vs 0.825.

\subsubsection{Performance with KD}

\begin{table*}[!t]
\begin{center} 
\caption{Quantitative comparisons between our method and other approaches built on large networks on the NYU-v2 dataset.}
\label{nyu_depth_acc}
\begin{tabular}{l|lcccc}
\hline
Method & Backbone & Params (M) $\downarrow$  & RMSE $\downarrow$ & REL $\downarrow$  & $\delta_1$ $\uparrow$ \\
\hline
Laina et al. \cite{laina2016deeper}&ResNet-50 &60.6 &0.573  &0.127  &0.811 \\ 
Hu et al. \cite{hu2019revisiting}&ResNet-50  &63.6 &0.555 &0.126  &0.843 \\ 
Zhang et al. \cite{Zhang2019PatternAffinitivePA} &ResNet-50  &95.4 &0.497 &0.121  &0.846  \\
Fu et al. \cite{fu2018deep}&ResNet-101  &110.0 &0.509 &0.115 &0.828 	 \\    

Hu et al.  \cite{hu2019revisiting}&SeNet-154  &149.8	&0.530 &0.115  &0.866 	 \\
Chen et al. \cite{Chen2019StructureAwareRP}&SeNet-154 &210.3 &0.514 & 0.111  & 0.878  \\ 
Chen et al.\cite{Chen2020ImprovingMD}&ResNet-101    &163.4 &\textbf{0.376} &\textbf{0.098} & \textbf{0.899} \\ \hline
Ours $N_s(\mathcal{X} \cup \mathcal{U})$ &MobileNet-V2 &1.7 &0.482 &0.131  &0.837\\ 
Ours $N_s(\mathcal{X} \cup \mathcal{U}^{'})$  &MobileNet-V2 &1.7 &\textbf{0.461} &\textbf{0.121}  &\textbf{0.855} \\ \hline
\end{tabular}
\end{center}
\vspace{-5mm}
\end{table*}

We conducted a series of experiments to validate our proposed method. We began by training the teacher networks $N_t(\mathcal{X})$ and $N_t(\mathcal{X} \cup \mathcal{U}^{'})$ on the datasets $\mathcal{X}$ and $\mathcal{X} \cup \mathcal{U}^{'}$, respectively. Subsequently, we trained the student network in four different settings:
\begin{enumerate}
    \item Using a trained teacher network on the original dataset, we applied knowledge distillation with the original training set, i.e., $N_t(\mathcal{X}) \rightarrow N_s(\mathcal{X})$.
    \item Using a trained teacher network on the original dataset, we applied knowledge distillation with the auxiliary unlabeled set, i.e., $N_t(\mathcal{X}) \rightarrow N_s(\mathcal{U})$.
    \item Using a trained teacher network on the original dataset, we applied knowledge distillation with both the original training set and the auxiliary unlabeled set, i.e., $N_t(\mathcal{X}) \rightarrow N_s(\mathcal{X}\cup \mathcal{U})$.
    \item Using a trained teacher network on both the original training set and auxiliary labeled set, we applied knowledge distillation with the mixed labeled set, i.e., $N_t(\mathcal{X}\cup \mathcal{U'}) \rightarrow N_s(\mathcal{X}\cup \mathcal{U'})$.
\end{enumerate}

The results in Fig.~\ref{w_wo_kd} (b) demonstrate a notable performance gap between the teacher and student networks when standard KD is applied in setting 1), with a drop in performance from 0.845 to 0.808. Interestingly, using only auxiliary unlabeled data in setting 2) leads to even better performance compared to standard KD. Combining the original training set and auxiliary unlabeled data in setting 3) results in a significant performance boost.

As shown in Fig.\ref{w_wo_kd} (a) for $N_s(\mathcal{X}\cup\mathcal{U'})$, when auxiliary data is labeled, the lightweight network's performance can be improved through supervised learning. However, due to the small network's limited capacity, the improvement is modest, and the network's performance is still inferior to that trained with KD and auxiliary unlabeled data, as seen in the result of $N_s(\mathcal{X}\cup\mathcal{U})$ in Fig.\ref{w_wo_kd} (b).

Moreover, a more accurate teacher can be learned to further improve the lightweight network's performance through KD, as seen in $N_s(\mathcal{X}\cup\mathcal{U'})$ of Fig.~\ref{w_wo_kd} (b).




\begin{figure*}[t]
\centering  
\begin{tabular}
{p{0.18\textwidth}<{\centering}p{0.18\textwidth}<{\centering}p{0.18\textwidth}<{\centering}p{0.18\textwidth}<{\centering}p{0.18\textwidth}<{\centering}} 
\IncG[ width=0.8in]{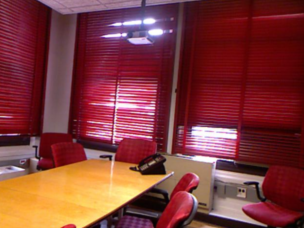}
&\IncG[ width=0.8in]{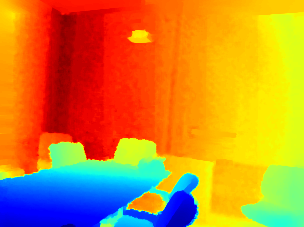}
&\IncG[ width=0.8in]{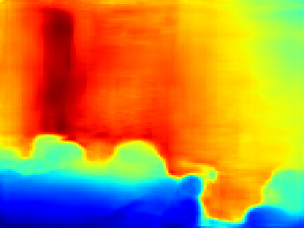}
&\IncG[ width=0.8in]{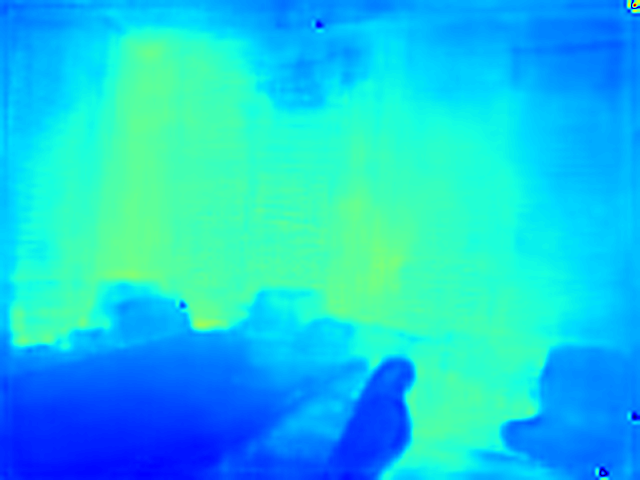}
&\IncG[ width=0.8in]{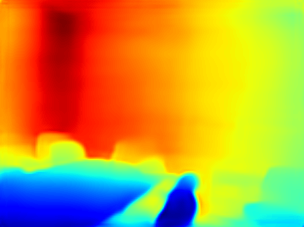}
\\
\IncG[ width=0.8in]{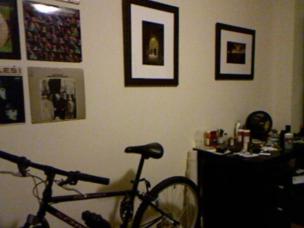}
&\IncG[ width=0.8in]{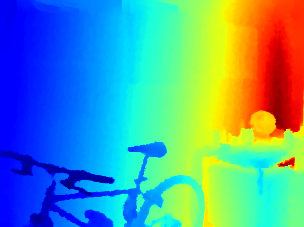}
&\IncG[ width=0.8in]{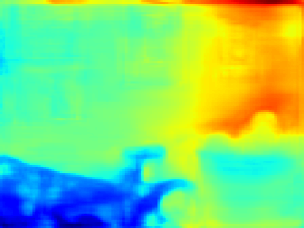}
&\IncG[ width=0.8in]{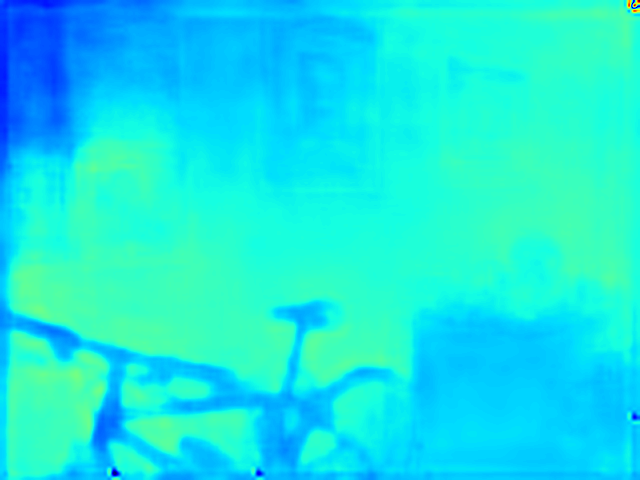}
&\IncG[ width=0.8in]{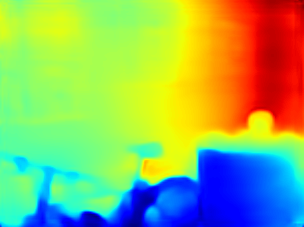}
\\
\IncG[ width=0.8in]{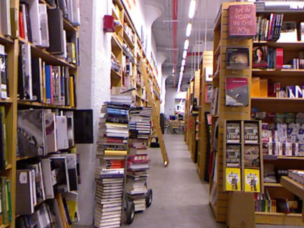}
&\IncG[ width=0.8in]{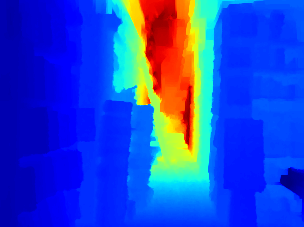}
&\IncG[ width=0.8in]{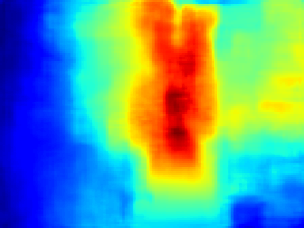}
&\IncG[ width=0.8in]{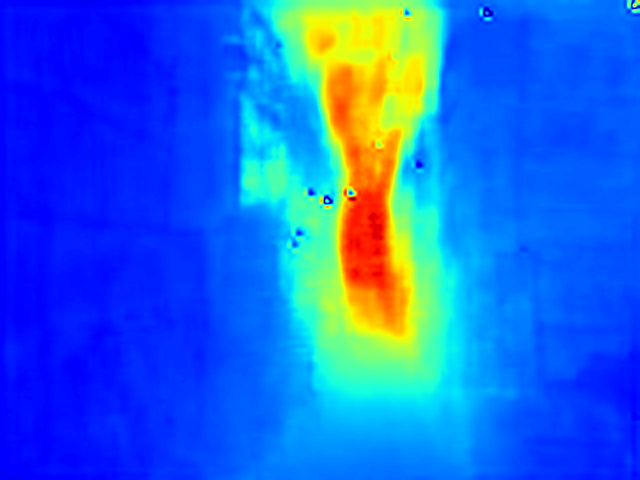}
&\IncG[ width=0.8in]{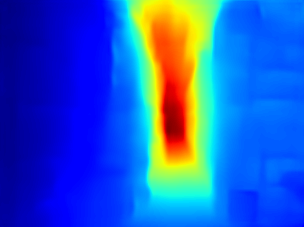}
\\
\footnotesize{(a) Input images} & \footnotesize{(b) Ground truth} & \footnotesize{(c) Fast-depth \cite{Wofk2019FastDepthFM}}& \footnotesize{(d) Joint-depth \cite{Nekrasov2019RealTimeJS}} &(e) \footnotesize{Ours} \\
\end{tabular}
\caption{Qualitative comparison of different methods for lightweight depth estimation on the NYU-v2 dataset. }
\label{fig1}
\end{figure*}


\subsection{Comparison with Previous Methods.}
\textbf{Comparison with Large Networks:}
Table~\ref{nyu_depth_acc} compares our method against previous methods built on different backbone networks, ranging from ResNet-50 to SeNet-154, demonstrating a clear trend of accuracy improvement. Notably, when utilizing only auxiliary unlabeled data, our method achieves comparable results to \cite{Zhang2019PatternAffinitivePA} and \cite{hu2019revisiting}, and even outperforms \cite{fu2018deep} and \cite{laina2016deeper} with a significantly smaller model size of only 1.7 M parameters.

In terms of methods utilizing extra labeled data, the best performance in Table~\ref{nyu_depth_acc} is achieved by \cite{Chen2020ImprovingMD}, where six auxiliary datasets with a total of 120K extra training data are carefully selected to handle hard cases for depth estimation, such as spurious edges and reflecting surfaces. While our method uses randomly selected auxiliary data from the ScanNet dataset, we believe that utilizing similar carefully selected data could further improve our method's performance.

\setlength{\tabcolsep}{1.0pt}
\begin{table}[!t]
\begin{center}
\caption{Quantitative comparison of lightweight approaches on the NYU-v2 dataset. The best and the second best results are highlighted in red and blue, respectively.}
\label{nyu_depth_acc_light}
\begin{tabular}{l|lccccccc}
\hline
Method & Backbone & Params (M) & GPU [ms]  & $\delta_1$  \\
\hline
Fast-depth \cite{Wofk2019FastDepthFM} & MobileNet &3.9 &7  &0.775\\
Joint-depth \cite{Nekrasov2019RealTimeJS} &MobileNet-V2 &3.1 &21 &0.790\\ 
Ours $N_s(\mathcal{X} \cup \mathcal{U})$     &MobileNet-V2 &1.7 &11   &{\color{blue}\textbf{0.837}}\\ 
Ours $N_s(\mathcal{X} \cup \mathcal{U}^{'})$   &MobileNet-V2 &1.7 &11    &{\color{red}\textbf{0.855}}\\ 
\hline
\end{tabular}
\end{center}
\vspace{-5mm}
\end{table}

\textbf{Comparison with lightweight Networks:}
We conducted a comparison between our proposed method and two previous approaches for lightweight depth estimation: Fast-depth \cite{Wofk2019FastDepthFM}, a traditional encoder-decoder net, and Joint-depth \cite{Nekrasov2019RealTimeJS}, which jointly learns semantic and depth information. Table~\ref{nyu_depth_acc_light} presents the quantitative results of this comparison, which show that our method outperforms the other two methods by a significant margin, even with only about half of the parameters. Specifically, the $\delta_1$ accuracy of our method, $N_s(\mathcal{X} \cup \mathcal{U})$, is $83.7\%$, which outperforms Joint-depth and Fast-depth by $4.7\%$ and $6.2\%$, respectively. Furthermore, when the auxiliary data is labeled, the improvement is more significant, as the accuracy of $N_s(\mathcal{X} \cup \mathcal{U'})$ is 85.5\%, representing $6.5\%$ and $8\%$ improvement over Joint-depth and Fast-depth, respectively. In addition, the qualitative comparisons in Fig.~\ref{fig1} show that the estimated depth maps of our method are more accurate and have finer details.

We also compared the GPU time required to infer a depth map from an input image. To conduct this comparison, we used a computer with an Intel(R) Xeon(R) CPU E5-2690 v3 and a GT1080Ti GPU card. We calculated the computation time for the other two methods using their official implementations. The results show that our method infers a depth map using only 11 ms of GPU time, which is much faster than Joint-depth. However, it is worth noting that Fast-depth achieves the smallest inference speed at the expense of degradation of accuracy and demonstrates the worst accuracy among the three methods.

\subsection{Effect of Varying the number of Auxiliary Data}
We conducted an ablation study to investigate the impact of varying the number of auxiliary data on the performance of our lightweight network. Specifically, we used the teacher model trained on the original labeled set and applied knowledge distillation with different numbers of unlabeled samples taken from $\mathcal{U}$. In our experiments, we evaluated our approach using 11.6K, 22.0K, 40.2K, 67.6K, 153.0K, and 204.2K auxiliary samples.

\begin{figure}[t]
\centering
\subfigure {\includegraphics[width=0.6\linewidth]{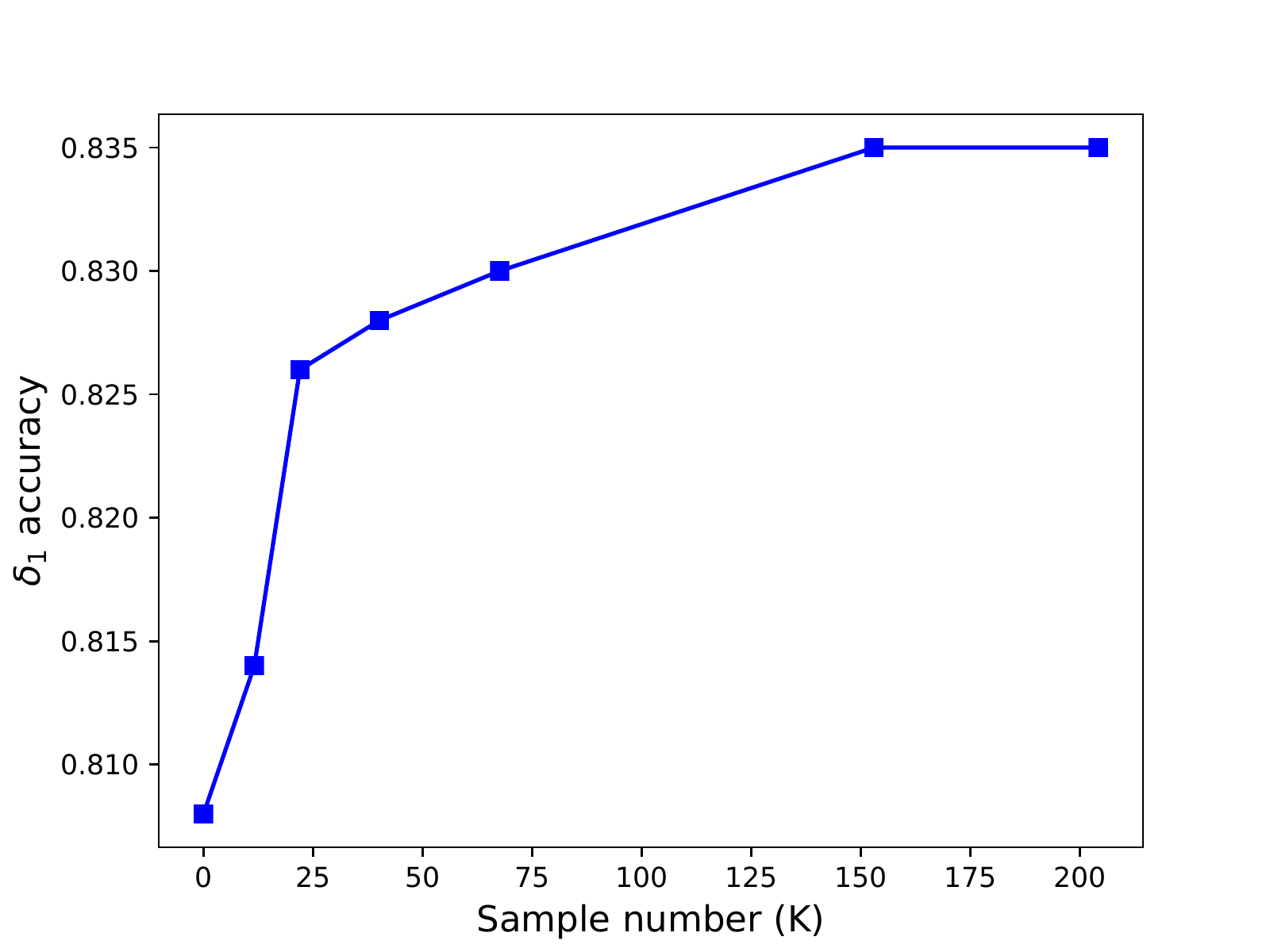}}
\vspace{-5mm}
\caption{Results of KD that applies  $N_t(\mathcal{X}) \rightarrow N_s(\mathcal{X \cup U})$ with different number of training samples from $\mathcal{U}$.}
\label{fig_num}
 \vspace{-5mm}
\end{figure}

As shown in Fig.~\ref{fig_num}, our results indicate that increasing the number of auxiliary data samples generally leads to better knowledge distillation performance. However, we observed diminishing returns after a certain number of samples, beyond which adding more samples did not yield any additional improvement.

\subsection{Cross-dataset Evaluation}
To assess the generalization performance of our lightweight model, we conduct a cross-dataset evaluation on two widely used datasets: SUNRGBD \cite{Song2015SUNRA} and TUM \cite{sturm12iros}. We directly apply our method, $N_s(\mathcal{X}\cup\mathcal{U})$ and $N_s(\mathcal{X}\cup\mathcal{U'})$, to evaluate on these datasets without any fine-tuning.
Note that the comparison between $N_s(\mathcal{X}\cup\mathcal{U'})$ and other methods may not be entirely fair as our method employs auxiliary labeled data. However, we include these results to demonstrate the effectiveness and reliability of utilizing auxiliary data to improve KD in a data-driven manner.
The results for each dataset are presented below.

\subsubsection{Results on SUNRGBD}
The generalization performance of our method is evaluated on the SUNRGBD dataset, which is commonly used in previous works for this purpose \cite{huynh2020guiding}. Table~\ref{sunrgbd_evaluation} presents the results, where the best and second best results are highlighted in red and blue, respectively. Our method achieves the lowest RMSE and REL error, while Joint-depth outperforms others in $\delta_1$ accuracy and ranks second in REL.


\begin{table}[t]
\begin{center}
\caption{The results of different methods on the SUNRGBD dataset. The best and the second best results are highlighted in red and blue, respectively.}
\label{sunrgbd_evaluation}
\begin{tabular}
{l|ccccccccccc}
\hline
Method  & RMSE && && & REL & && && $\delta_1$\\ \hline
Fast-depth \cite{Wofk2019FastDepthFM}  &0.662  &&&&&0.376 &&& &&0.404 \\
Joint-depth \cite{Nekrasov2019RealTimeJS} &0.634&&  &&&{\color{blue} \textbf{0.338}}  &&&&&{\color{red} \textbf{0.454}} \\
Ours $N_s(\mathcal{X} \cup \mathcal{U})$ & {\color{blue} \textbf{0.577}} &&& & &{\color{blue} \textbf{0.338}}& & && &0.430\\
Ours $N_s(\mathcal{X} \cup \mathcal{U'})$  &{\color{red} \textbf{0.531}}&&  &&&{\color{red} \textbf{0.306}} &  &&&&{\color{blue} \textbf{0.446}}\\
\hline
\end{tabular}
\end{center}
\vspace{-5mm}
\end{table}

\begin{table}[t]
\begin{center}
\caption{The $\delta_1$ accuracy of different methods on the five sequences from TUM dataset. The best and the second best results are highlighted in red and blue, respectively.}
\label{cross_evaluation}
\begin{tabular}
{l|ccccc}
\hline
Method &360 &desk &desk2 &rpy &xyz \\ \hline
Fast-depth \cite{Wofk2019FastDepthFM}   &0.548 &0.308 &0.358 &0.333 &0.287 \\
Joint-depth \cite{Nekrasov2019RealTimeJS} &0.512 &0.410 &0.441 &0.552 &{\color{blue} \textbf{0.583}}\\
Ours $N_s(\mathcal{X} \cup \mathcal{U})$  &{\color{blue} \textbf{0.615}}  &{\color{blue} \textbf{0.442}} &{\color{blue}\textbf{0.498}} &{\color{blue}\textbf{ 0.611}} &0.486 \\
Ours $N_s(\mathcal{X} \cup \mathcal{U'})$  &{\color{red} \textbf{0.854}} &{\color{red} \textbf{0.695}} &{\color{red} \textbf{0.772}}  &{\color{red} \textbf{0.679}} &{\color{red} \textbf{0.905}}\\
\hline
\end{tabular}
\end{center}
\vspace{-5mm}
\end{table}

\subsubsection{Results on TUM}
We evaluated the generalization performance of our lightweight depth estimation method on the TUM dataset using five sequences, namely fr1/360, fr1/desk, fr1/desk2, fr1/rpy, and fr1/xyz, as in \cite{Czarnowski2020DeepFactorsRP}. Depth accuracy was measured by $\delta_1$. As shown in Table~\ref{cross_evaluation}, our method significantly outperforms the other methods, demonstrating a satisfactory generalization performance.

The average $\delta_1$ accuracy for Joint-depth and Fast-depth is 0.494 and 0.369, respectively, while our method with only auxiliary unlabeled data, $N_s(\mathcal{X} \cup \mathcal{U})$, achieves an average accuracy of 0.530. When auxiliary labeled data is used, our method $N_s(\mathcal{X} \cup \mathcal{U'})$ achieves an even higher accuracy, with an average of 0.781.


\section{Conclusion}
n this paper, we revisit the problem of monocular depth estimation by focusing on the balance between inference accuracy and computational efficiency. We identify the inherent challenge of striking a balance between accuracy and model size. To address this challenge, our method proposes a lightweight network architecture that significantly reduces the number of parameters. We then demonstrate that incorporating auxiliary training data with similar scene scales is an effective strategy for enhancing the performance of the lightweight network. We conduct two experiments, one with auxiliary unlabeled data and one with auxiliary labeled data, both utilizing knowledge distillation. Our method achieves comparable performance with state-of-the-art methods built on much larger networks, with only about 1\% of the parameters, and outperforms other lightweight methods by a significant margin.

\vspace{-3mm}
\subsubsection{Acknowledgements} This work was partly supported by the National Natural Science Foundation of China (62073274, 62106156), Shenzhen Science and Technology Program (JCYJ20220818103000001), and the funding AC01202101103 from the Shenzhen Institute of Artificial Intelligence and Robotics for Society.
\vspace{-3mm}

%
%
%
\bibliographystyle{splncs04}
\bibliography{egbib}

\end{document}